\documentclass{article}
\pdfoutput=1

\usepackage{arxiv}

\usepackage[utf8]{inputenc} 
\usepackage[T1]{fontenc}    
\usepackage{natbib}
\usepackage{url}            
\usepackage{booktabs}       
\usepackage{amsfonts}       
\usepackage{nicefrac}       
\usepackage{microtype}      
\usepackage{lipsum}		
\usepackage{graphicx}
\usepackage{subfig}
\usepackage{doi}
\usepackage{amsmath,amssymb,amsfonts}

\title{Exploring the Properties and Evolution of Neural Network Eigenspaces during Training} 

\author{Mats L. Richter \\
    Department of Cognitive Science\\
	University of Osnabrück\\
	49080, Osnabrück \\
	\texttt{matrichter@uni-osnabrueck.de} \\
	\And
	Leila Malihi \\
    Department of Cognitive Science\\
	University of Osnabrück\\
	49080, Osnabrück \\
	\texttt{lmalihi@uni-osnabrueck.de} \\
	\And
	Anne-Kathrin Patricia Windler \\
    Department of Cognitive Science\\
	University of Osnabrück\\
	49080, Osnabrück \\
	\texttt{awindler@uni-osnabrueck.de} \\
	\And
	Ulf Krumnack \\
    Department of Cognitive Science\\
	University of Osnabrück\\
	49080, Osnabrück \\
	\texttt{krumnack@uni-osnabrueck.de} \\
}

\renewcommand{\undertitle}{Preprint (Under Review)}

\hypersetup{
pdftitle={Exploring the Properties and Evolution of Neural Network Eigenspaces during Training},
pdfsubject={q-bio.NC, q-bio.QM},
pdfauthor={Mats L. Richter, Leila Malihi, Ulf Krumnack},
pdfkeywords={convolutional neural networks, logistic regression probes, saturation},
}

\begin{document}
\maketitle

\begin{abstract}
In this work we explore the information processing inside neural networks using logistic regression probes \cite{probes} and the saturation metric \cite{featurespace_saturation}.
We show that problem difficulty and neural network capacity affect the predictive performance in an antagonistic manner, opening the possibility of detecting over- and under-parameterization of neural networks for a given task.
We further show that the observed effects are independent from previously reported pathological patterns like the ``tail pattern'' described in \cite{featurespace_saturation}. Finally we are able to show that saturation patterns converge early during training, allowing for a quicker cycle time during analysis 
\end{abstract}

\keywords{convolutional neural networks \and logistic regression probes \and saturation}

\section{Introduction}
Deep Convolutional Neural Networks are opaque machine learning solutions.
Opaque in the sense that the model state itself is neither self evident nor human interpretable.
This has lead to a primarily trial and error driven approach for development, that relies on the comparison of abstract, model agnostic performance metrics like accuracy as a measure for classification performance, the number of parameters as a measure of capacity and the FLOPs per forward pass as a metric for computational efficiency \cite{nasnet, efficientnet, resnet}.
To move towards a more efficient, less trial and error based design process, a deeper understanding of the model's state is required.
This understanding does not have to be necessarily complete with regards to fully understanding the relation of the input and output of the model.
The comparative analysis methods based on SVCCA \cite{svcca} are good examples of such a non-holistic approaches.
The information extracted from the model by using SVCCA is highly aggregated but allows for useful insights into the converged model.
Logistic Regression Probes \cite{probes} and saturation \cite{featurespace_saturation} aggregate a single layer to a number, which allows for easy and intuitive analysis, similar to measuring with a thermometer.
While logistic regression probes measure the intermediate solution quality very directly by training logistic regressions on the output of a layer, saturation is more task agnostic.
The authors show that the size of the subspace of the feature space \footnote{We refer to feature space as the space in which the data exists in a specific layer`s output.} responsible for data processing varies significantly depending on the input resolution, leading to inefficiencies.
In this work we explore the properties of saturation further by answering the following questions:

\begin{itemize}
    \item How is saturation influenced by model capacity and problem difficulty?\footnote{We define capacity of a network by the number of trainable parameters, which is also common in other works \cite{efficientnet, amoebanet, gpipe}.}?
    Answer: Yes, we show that problem difficulty and model capacity behave inversely proportional to each other in their influence on saturation.
    \item Does the capacity (number of filters) of individual layers influence how the inference process is distributed over the network?
    Answer: No, the tested models seem to be unable to shift processing to other layers.
    \item How do saturation patterns evolve during training?
    Answer: They converge at a similar pace as the loss of the model, with saturation increasing during training.
    The way saturation evolves also gives hints on the properties of the dataset, but it is  not influenced by the model over-fitting.
\end{itemize}

\section{Background}

\subsection{Logistic Regression Probes}
For the analysis of trained models, we use probe classifiers.
Probe classifiers are a tool for analyzing how the solution quality progresses while the data is propagated through the network's structure \cite{probes}.
To do this, logistic regression "probes"  are trained on the same task as the model, using the output of individual layers as input.
Since the softmax layer and the probes effectively solve the same task, the probes can be used to judge the quality of the intermediate solutions.
Typically, probe performance increases layer by layer, approaching the model's performance (as figure~\ref{fig:vgg16_cifar10} demonstrates).
We refer to the test accuracy of a probe computed on the output of layer $l$ as $p_l$

\subsection{Saturation}
The saturation $s_l$ of a layer $l$ is a simple scalar metric that can be applied to any layer in a neural network and was first introduced by \cite{spectral_analysis} an explored in more detail by \cite{featurespace_saturation}.
In simple terms, saturation is a percentage value that measures how many of the available dimensions in the output space $Z_l$ of the layer $l$ are relevant for the inference process.
\begin{equation}
    s_l = \frac{\operatorname{dim} E^k_l}{\operatorname{dim} Z_l}
\end{equation}

Saturation is computed by approximating the ratio of the dimensionality of the relevant eigenspace $\operatorname{dim} E^k_l$ of layer $l$ and the extrinsic dimensionality of layer $l$s activation values $\operatorname{dim} Z_l$.
The relevant eigenspace $E^k_l$ is a subspace of $Z_l$ in which the information is processed.
We refer to this space as "relevant", since a projection of the data into the relevant eigenspace after the layers output will not result in a loss of predictive performance \cite{featurespace_saturation}.
The relevant eigenspace can be considered the subspace in which the information processing is happening.
The approximation of $E^k_l$ is done using PCA, where the largest eigendirections are kept in order to explain $99\%$ of the data`s variance in the output of layer $l$.
Because of this computation strategy, saturation can be computed on-line during training.
This is in contrast to logistic regression probes, which generally require the extraction of all latent representations and training of logistic regression probes, which may take longer than the actual training itself.

\subsection{The Semantics of Saturation}
\begin{figure}[tb!]
	\centering
	\includegraphics[width=0.9\columnwidth]{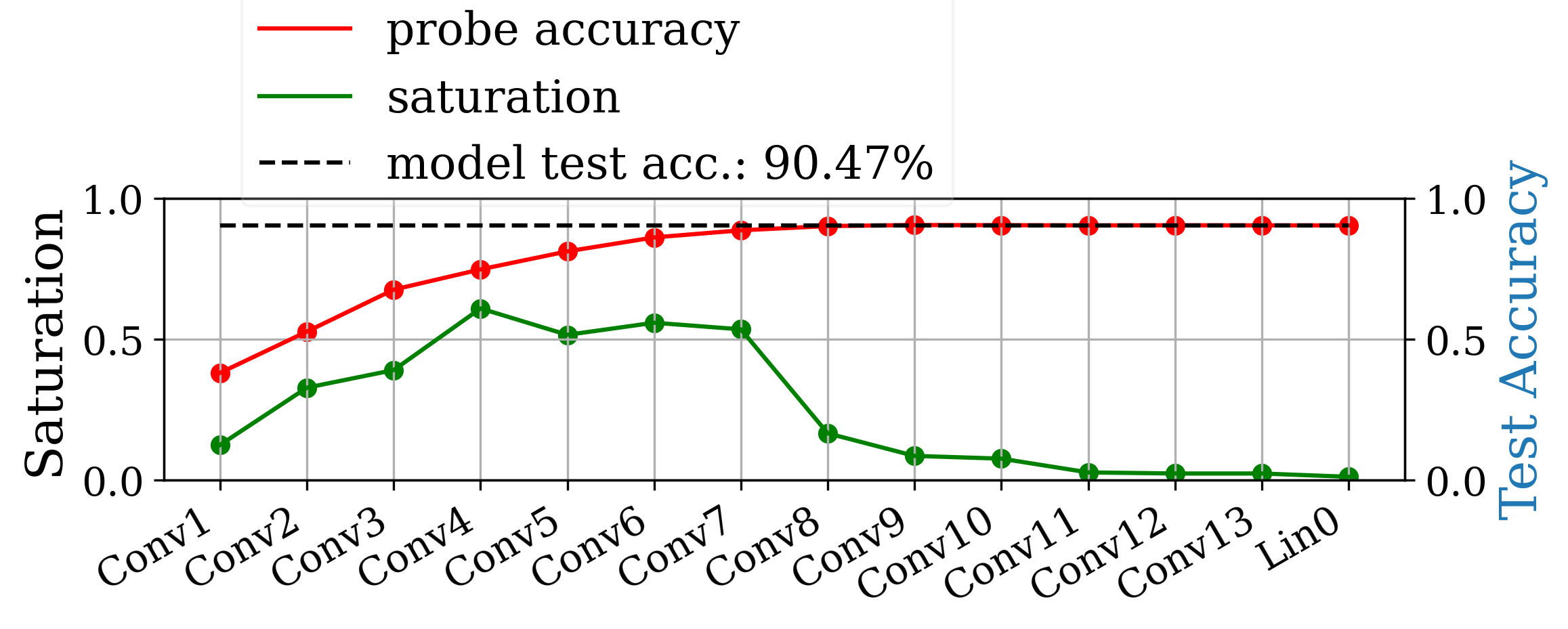}
	\caption{VGG16 exhibits a tail pattern starting from Conv8. In the tail, saturation is low and probe accuracy is stagnating at the same level as the model output.}
	\label{fig:vgg16_cifar10}
\end{figure}

A sequence of low saturated layers ($<$ 50\% of the average saturation of all other layers) is referred to as a ``tail pattern'' and indicates that these layers are not contributing qualitatively to the prediction (see figure \ref{fig:vgg16_cifar10}).
This is also visible when observing the performance of the probes, which stagnates in the tail-layers.

From these results, we can see that solving a problem saturates the layer more than simply passing through information in a given layer.
However, this does not mean that the absolute saturation value is indicative of the activity within a layer.
So far, saturation has been only explored as a quicker on-line computable analogy for logistic regression probes.
As such saturation has been viewed always relative to other layers within a neural network.
In this work, we will explore how the absolute saturation value changes in different scenarios.
We further explore how the saturation evolves during training and we will explain for the low dimensionality of the tail pattern.

\section{Capacity and Problem Difficulty behave proportionally}
In this section we analyze the relationship between problem difficulty and model capacity by addressing two questions exploring how this relationship is reflected in the saturation values.
In order to do so, we train the entire VGG-network family (VGG11, 13, 16 and, 19) on Cifar10 \cite{cifar} and reduce their capacity evenly over the entire architecture to observe how this behavior affects saturation values.
Our first hypothesis states that the average saturation $s_{\mu}$ increases proportionally with a reduction in capacity and a decrease in performance.
We then move on to investigate how the problem difficulty of the network changes the saturation emerging in a neural architecture.
Since the relevant eigenspace is generally larger when the layer is contributing to the quality of the solution \cite{featurespace_saturation}, we further hypothesize that more processing in a layer requires a larger relevant eigenspace.
If this assumption holds true, the overall saturation level should increase with an increase in the difficulty of the task.
If both working hypotheses are true, we can conclude that the difficulty of the problem and the capacity of the layers influence saturation in an antagonistic way.

\subsection{Methodology}
We test our working hypotheses by conducting two experiments.
We first train the VGG-family \cite{vgg} of networks on Cifar10.
We further train 4 additional variants of each model, which have the respective number of filters (and thus capacity) reduced by a factor of $\frac{1}{2}$, $\frac{1}{4}$, $\frac{1}{8}$ and $\frac{1}{16}$.
We choose Cifar10 for its manageable size, which allows for a larger number of model training runs to be conducted with our available resources, which is necessary for this experiment.
We choose the VGG-family of networks for its architectural simplicity and because we can test different depths of convolutional neural networks by experimenting on the entire family of networks.
Training itself is conducted using a stochastic gradient descent (SGD) optimizer with a learning rate of 0.1, which is decaying after 10 epochs with a decay factor of 0.1.
The models are trained on a batch size of 64 for 30 epochs in total.

The second experiment is conducted on ResNet18. We are using a similar setup to the first set of experiments. However, we are using a standardized input resolution of $224 \times 224$ pixels, to avoid artifacts caused by the input resolution.
We train the model on multiple datasets of different difficulties (in ascending order of complexity): MNIST, Cifar10, TinyImageNet, and the ImageNet dataset \cite{mnist, cifar, tinyimnet, imagenet}.
While it is hard to precisely define the complexity of the task, we think the selected datasets can be regarded as increasingly difficult based on the number of classes and the complexity of the visual information provided as data points to the model.
MNIST features strictly binary images of $28 \times 28$ pixel native resolution for a 10-class classification problem.
Cifar10 features $32 \times 32$ real world RGB images belonging to 10 classes.
TinyImageNet consists of $64 \times 64$ images with 200 classes and ImageNet is made up of RGB images of various sizes belonging to 1.000 classes.

\subsection{Results}
\begin{figure}[htb!]
	\centering
	\includegraphics[width=0.8\columnwidth]{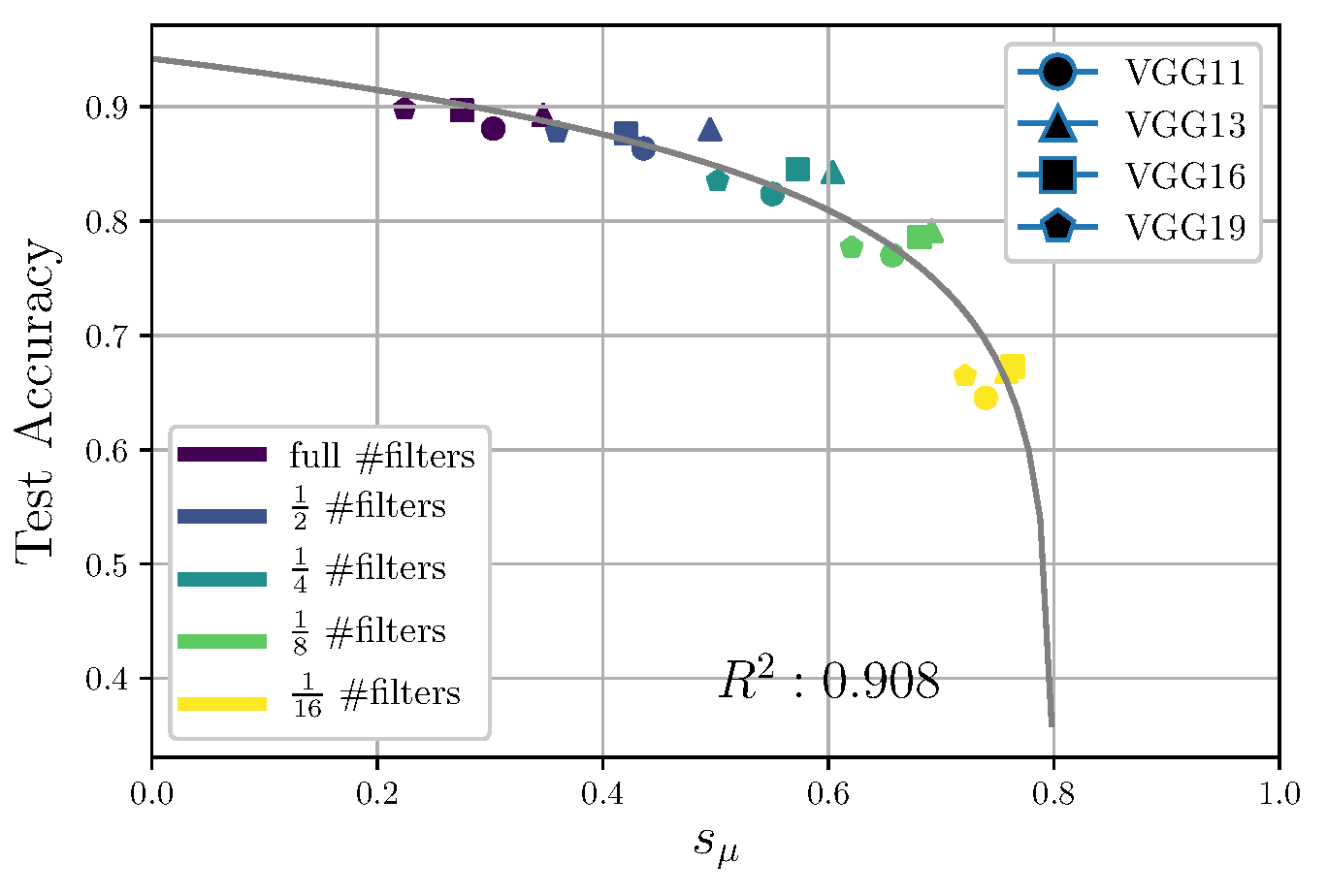}
	\includegraphics[width=0.8\columnwidth]{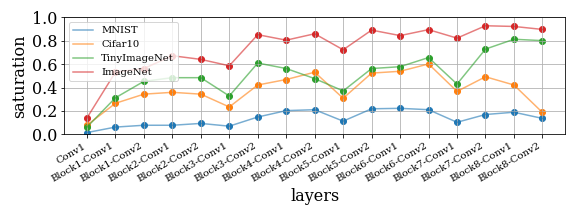}
	\caption{Reducing the model capacity by reducing the number of filters increases the average saturation and decreases performance (upper plot). Training a model on more difficult datasets also increases the overall saturation level. This indicates that saturation can measure the load on a  ResNet18 model (lower plot).}
	\label{fig:filters_avg_sat}
\end{figure}

The average saturation $s_{\mu}$ is increasing and the predictive performance is reduced when the capacity of the model is reduced.
The exponential reduction in capacity is reflected in a logarithmic relation between the increasing $s_{\mu}$ and predictive accuracy measured on the test set (see figure \ref{fig:filters_avg_sat}).
From these observations, we can conclude that reducing the capacity of the architecture leads to an increase in saturation.

We further observe in figure \ref{fig:filters_avg_sat} that  saturation also increases with problem complexity.
The saturation levels of all layers increase when the model is trained on a more difficult problem.
The overall shape of the saturation curve only deviates slightly, with no tail pattern or similar anomalous shapes emerging.
Since we know from the works of \cite{featurespace_saturation} that a resolution of $224 \times 224$ pixels results in an even distribution of the inference process for the trained model, we can conclude that less processing is required for less complex problems.

Combined with the insights gained from training the VGG-variants on Cifar10, we can conclude that for the pairs of dataset and model in our experiments a saturation ``sweet spot'' exists between $s_{\mu}$ = 0.2 and $s_{\mu} = 0.4$, which yields the good predictive performance without being too excessively overparameterized.

\section{On the emergence of saturation patterns}
\label{sec:emergence}
The tail pattern that we discussed earlier in this work allows for the identification of inefficiencies caused by mismatches between the neural architecture and the input resolution.
The analysis of saturation patterns is generally done after training has concluded.
However, since saturation can be computed life during training with little overhead \cite{featurespace_saturation}, we think that it might be interesting to see how these patterns emerge during the training process.

\subsection{Methodology}
We first investigate how the saturation level evolves in a single layer under different circumstances.
We train a 3 layer fully connected neural network, the first layer has 256 units, the second layers has 8, 16, 32, 64 and 128 units (a different number for each run).
We train these networks using the ADAM optimizer and a batch size of 128.
The training is conducted twice. Once using 8 epochs, which is enough for all models to converge, whereas the second experiment is run for 20 epochs, which results in the loss increasing again due to overfitting.
We compute the saturation of the hidden layer of the 3-layer architecture after each epoch to observe the evolution of the architecture.
We further compute the loss of the architecture to observe a potential relationship between loss and saturation convergence.

Based on these observations, we repeat the experiment on VGG11 and 19 as well as sparse (low capacity) versions of these models with $\frac{1}{8}$ of the original number of filters.
We do this to observe whether the evolution of saturation patterns is also depending on the architecture, depth, and capacity of the network.

\subsection{Results}
\begin{figure*}[htb!]
	\centering
	\subfloat[Validation loss during training.]{
	    \includegraphics[width=0.4\columnwidth]{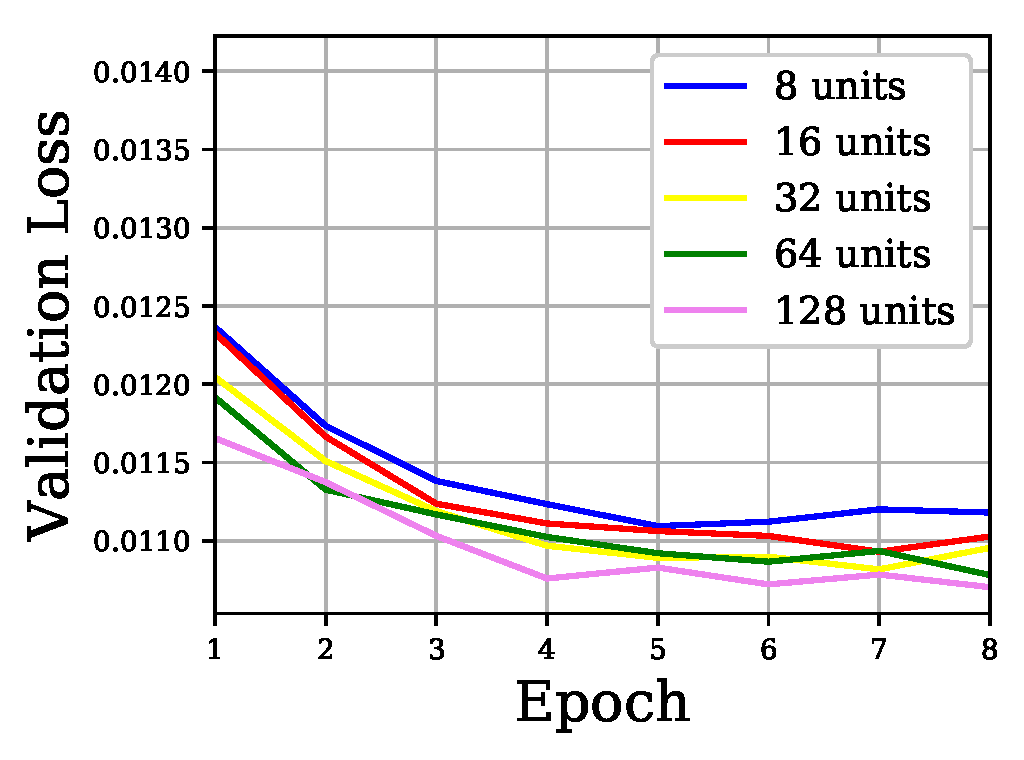}
	}
	\subfloat[Saturation of layer 2 during training.]{
	    \includegraphics[width=0.4\columnwidth]{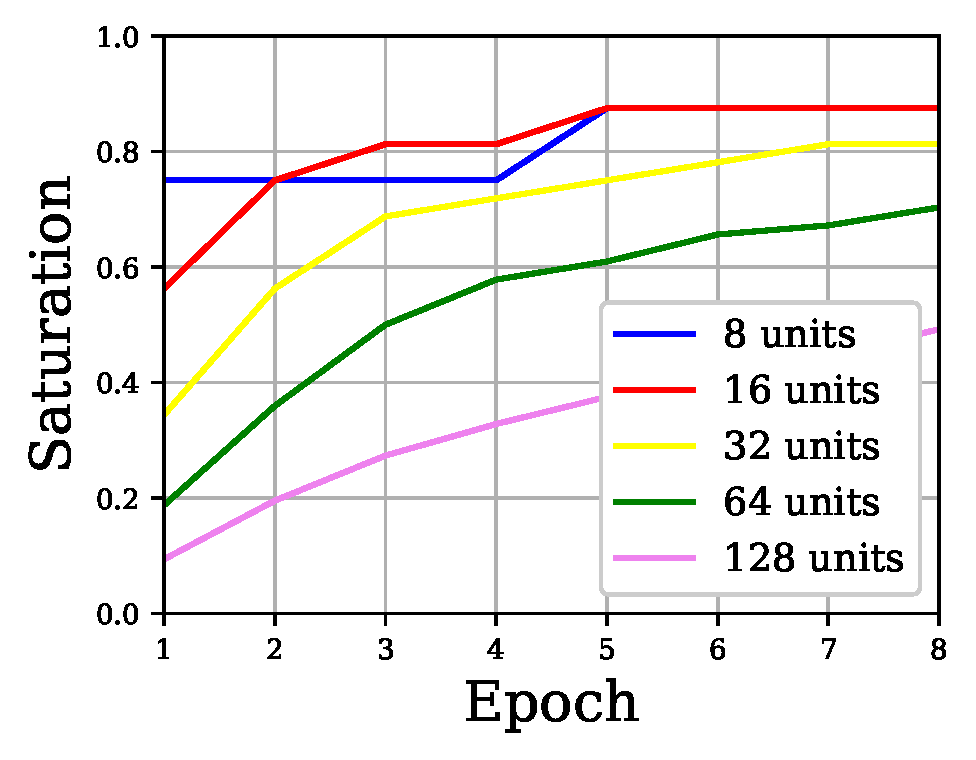}
	}
	\caption{Saturation and loss converge at a similar pace.}
    \label{fig:learning_curves_dense}
\end{figure*}

In figure \ref{fig:learning_curves_dense}, we observe that an increase in the number of units in the fully connected layer will result in an increased saturation during every epoch.
Furthermore, the saturation converges at a similar pace than the loss, starting from a low-saturation level and increasing epoch by epoch.

\begin{figure*}[ht!]
	\centering
	\subfloat[Validation loss during training.]{
	    \includegraphics[width=0.4\columnwidth]{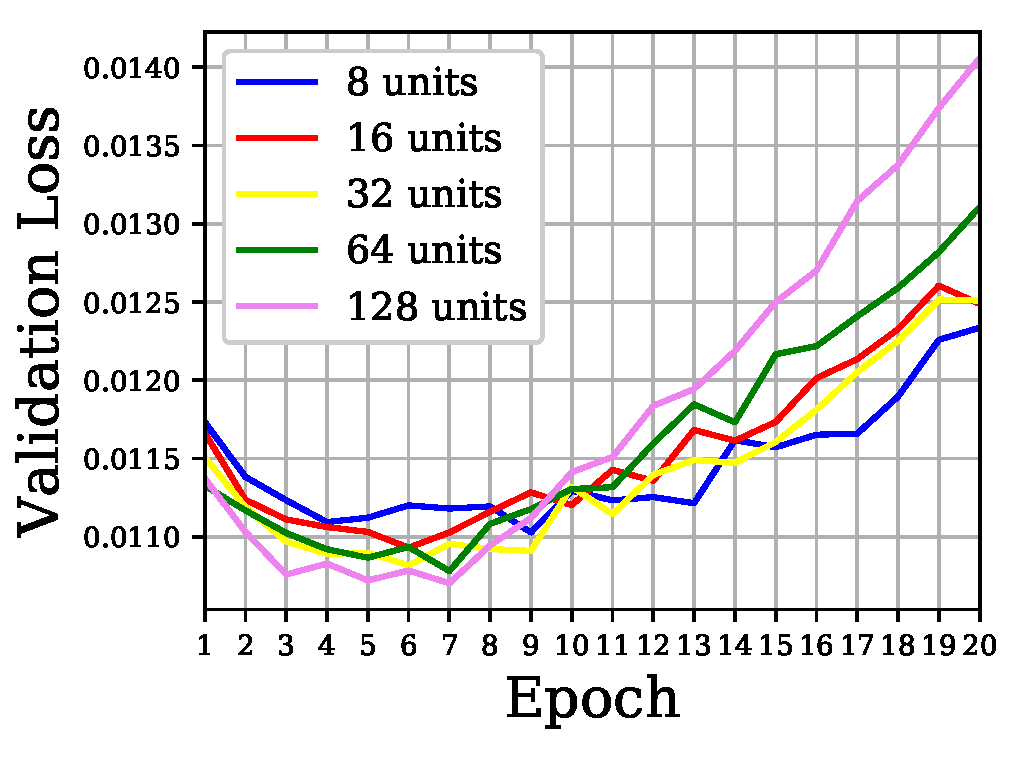}
	}
	\subfloat[Saturation of layer 2 during training.]{
	    \includegraphics[width=0.4\columnwidth]{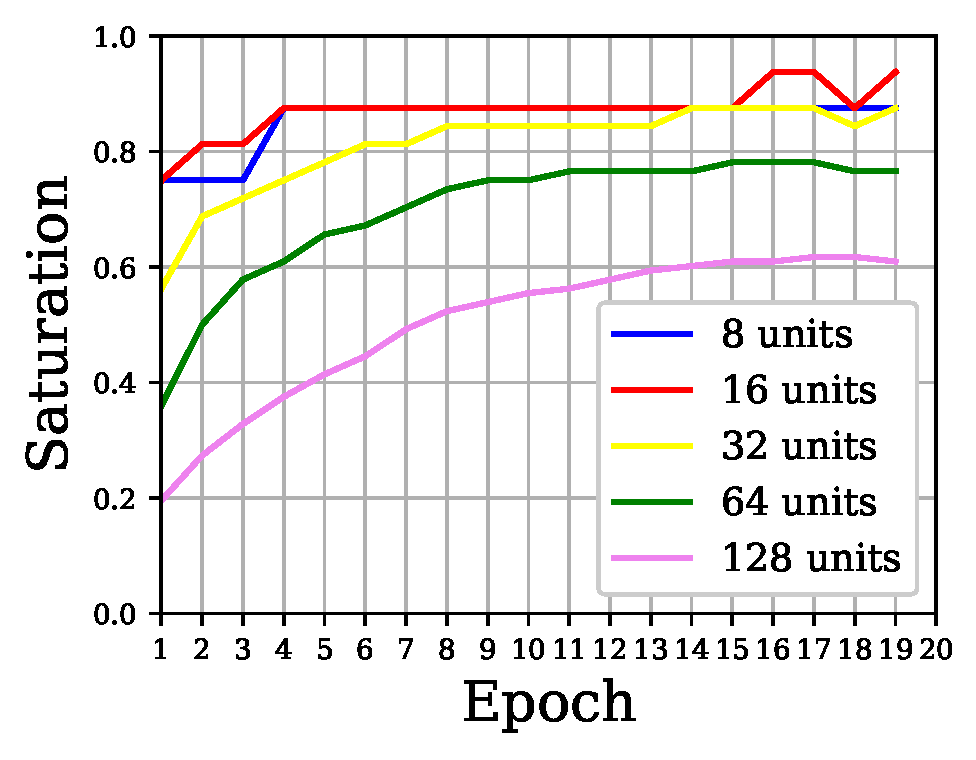}
	}
	\caption{When overfitting the saturation pattern keeps converging, indicating that overfitting is not drastically altering the relevant eigenspace.}
    \label{fig:overfit_loss}
\end{figure*}

This joint convergence is interrupted when the model starts to overfit.
In figure \ref{fig:overfit_loss}, we can see that the increase of validation loss is not followed by a drastic change in saturation. Instead, the curve of saturation values flattens for all sizes of the second layer.
The fact that overfitting is not reflected in saturation values indicates that the changes to the way the data is processed when the model starts to overfit are subtle and thus are not reflected in changes to the relevant eigenspace and therefore saturation.

\begin{figure}[tb!]
	\centering
	\label{fig:saturations}
	\includegraphics[width=0.6\columnwidth]{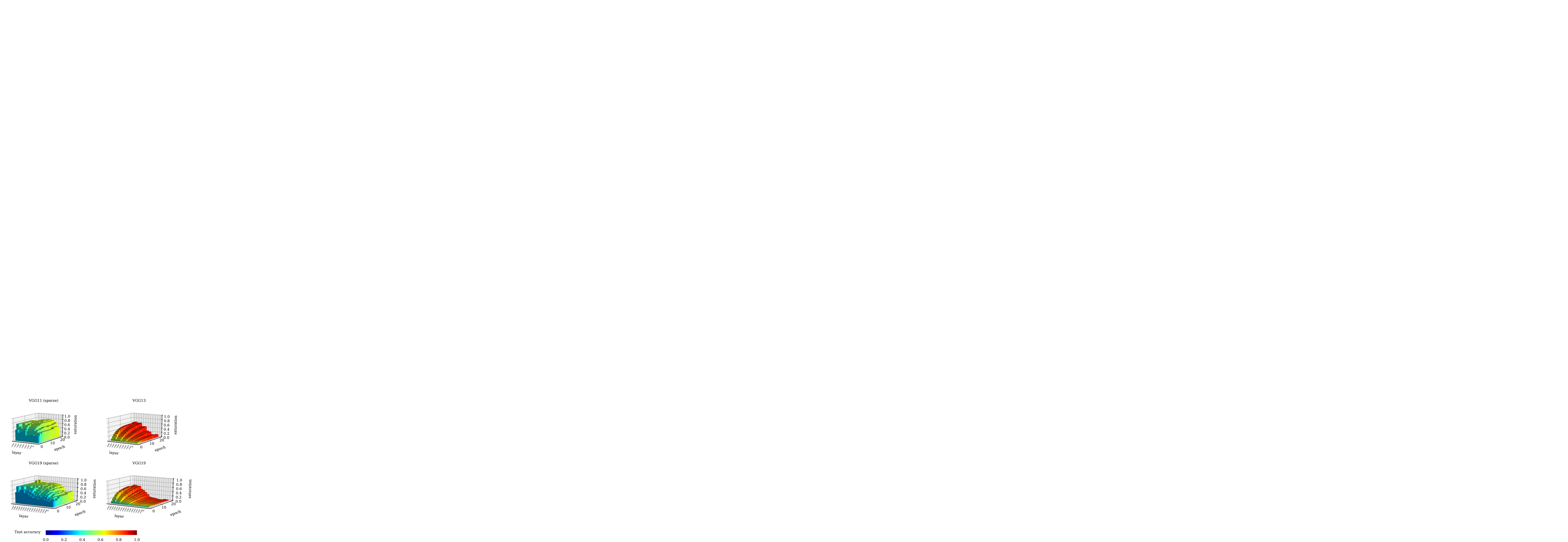}
	\caption{Saturations of convolutional neural networks show a converging behavior regarding saturation similar to previous observations in figure \ref{fig:learning_curves_dense}.}
	\label{fig:3d}
\end{figure}

In figure~\ref{fig:3d}, we can see that the converging behavior observed with the fully connected network can also be observed in a fully convolutional network.
This converging behavior is independent of the position of the layer in the network, the number of layers, and the capacity of the network.

Another interesting observation is that tail pattern seems to be observable rather early during training, which indicates that an on-line analysis during training allows the data scientist to detect inefficiencies early before training has concluded.

\section{Predictability of Tail Patterns Regarding Complexity}
In the previous section, we show that the overall saturation increases when the network's capacity is low.
In this section, we investigate how this affects the predictability of tail patterns.
\cite{sizematters} show that the tail patterns in sequential convolutional neural networks can be predicted by computing the receptive field of all convolutional layers.
The receptive field size is the height and width of the area on the input image that can influence the value of a single unit in a convolutional activation map.
The first layers ingesting an input with a receptive field size greater than the input resolution are referred to as the ``border layers'' and mark the start of the tail pattern.
In section~\ref{sec:emergence}, we have seen that the global saturation level can be altered by changing the number of filters in a convolutional layer.
Since technically the required capacity is no longer present in a single layer, the network could distribute the processing among additional layers in the architecture to process the data.
However, if the receptive field expansion is determining the number of unproductive layers (like the authors of \cite{sizematters} suggest), we will observe a tail pattern of unproductive layers starting at the border layer, which is the same layer as in the full capacity networks.

\subsection{Methodology}
We test the hypothesis by repeating the experiments conducted by \cite{sizematters} regarding the prediction of unproductive layers.
The authors of \cite{sizematters} were able to predict unproductive layers by computing the border layer for VGG11, 13, 16, and 19 on Cifar10.
We reduce the capacity of these models by reducing the filter size to $\frac{1}{8}$ of the original size to see whether a drastic loss in capacity changes how the inference is distributed.
The models are trained for 30 epochs using the SGD-optimizer with a learning rate of 0.1, decaying by a factor of 0.1 every 10 epochs. 
The batch size is 64, each batch is channel-wise normalized, each image is randomly cropped during inference time as well as randomly horizontally flipped with a probability of 50\%.
The receptive field and the border layer are computed using the formulas provided by \cite{sizematters}.

\subsection{Results}

\begin{figure}[t!]
	\centering
	\subfloat[VGG11]{
	    \includegraphics[width=0.4\columnwidth]{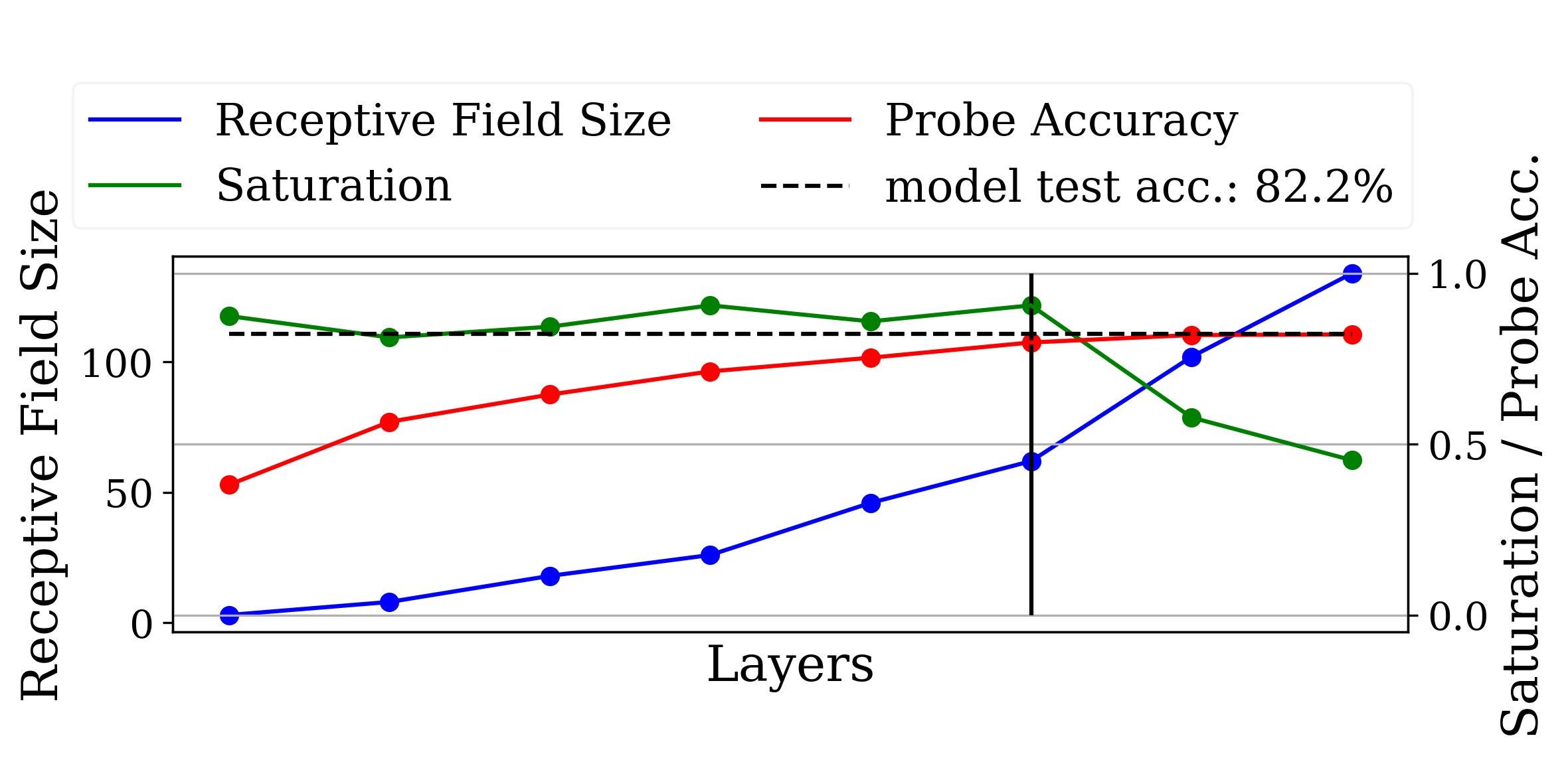}
	}
	\subfloat[VGG13]{
	    \includegraphics[width=0.4\columnwidth]{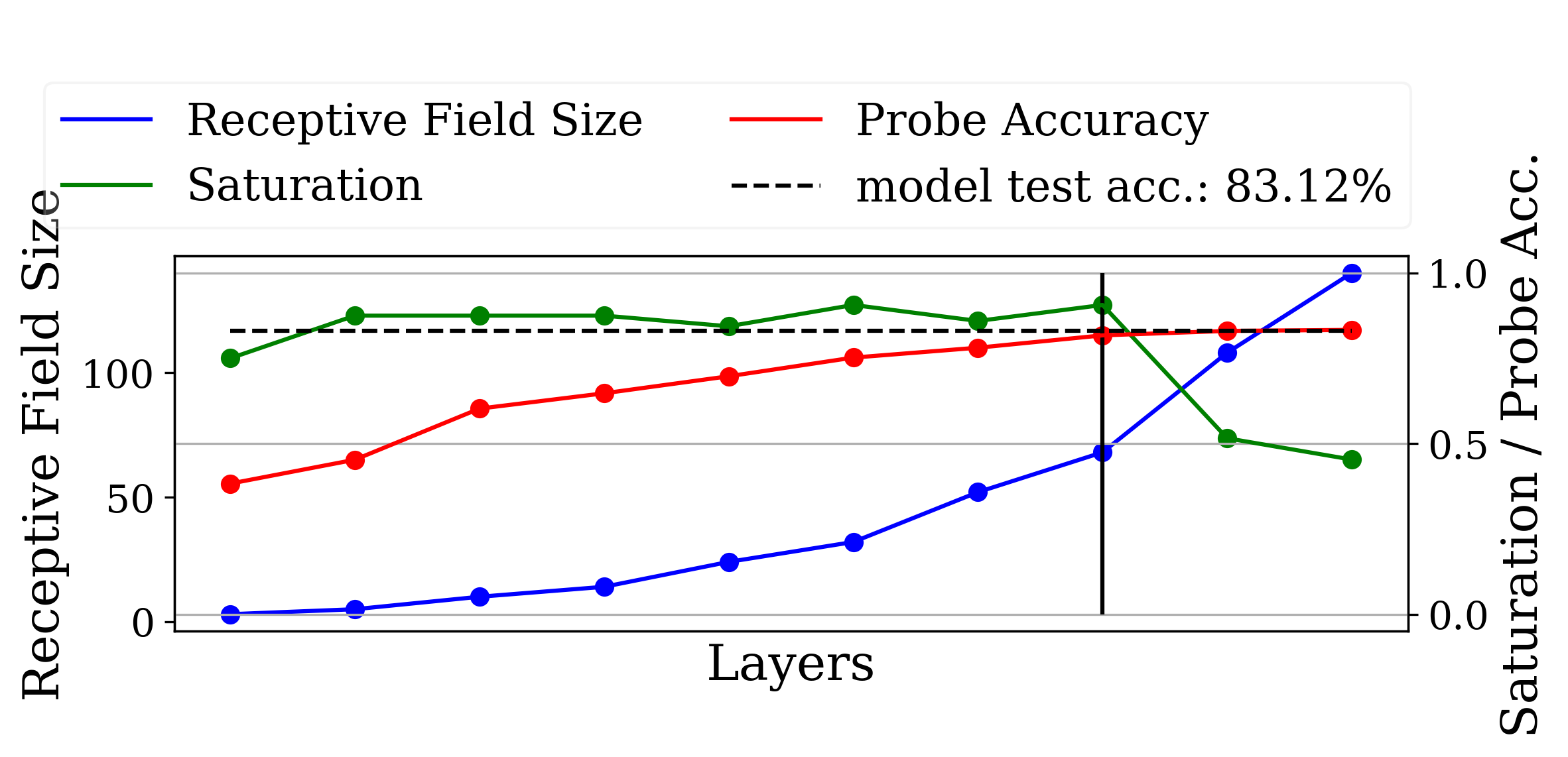}
	}\qquad
		\subfloat[VGG16]{
	    \includegraphics[width=0.4\columnwidth]{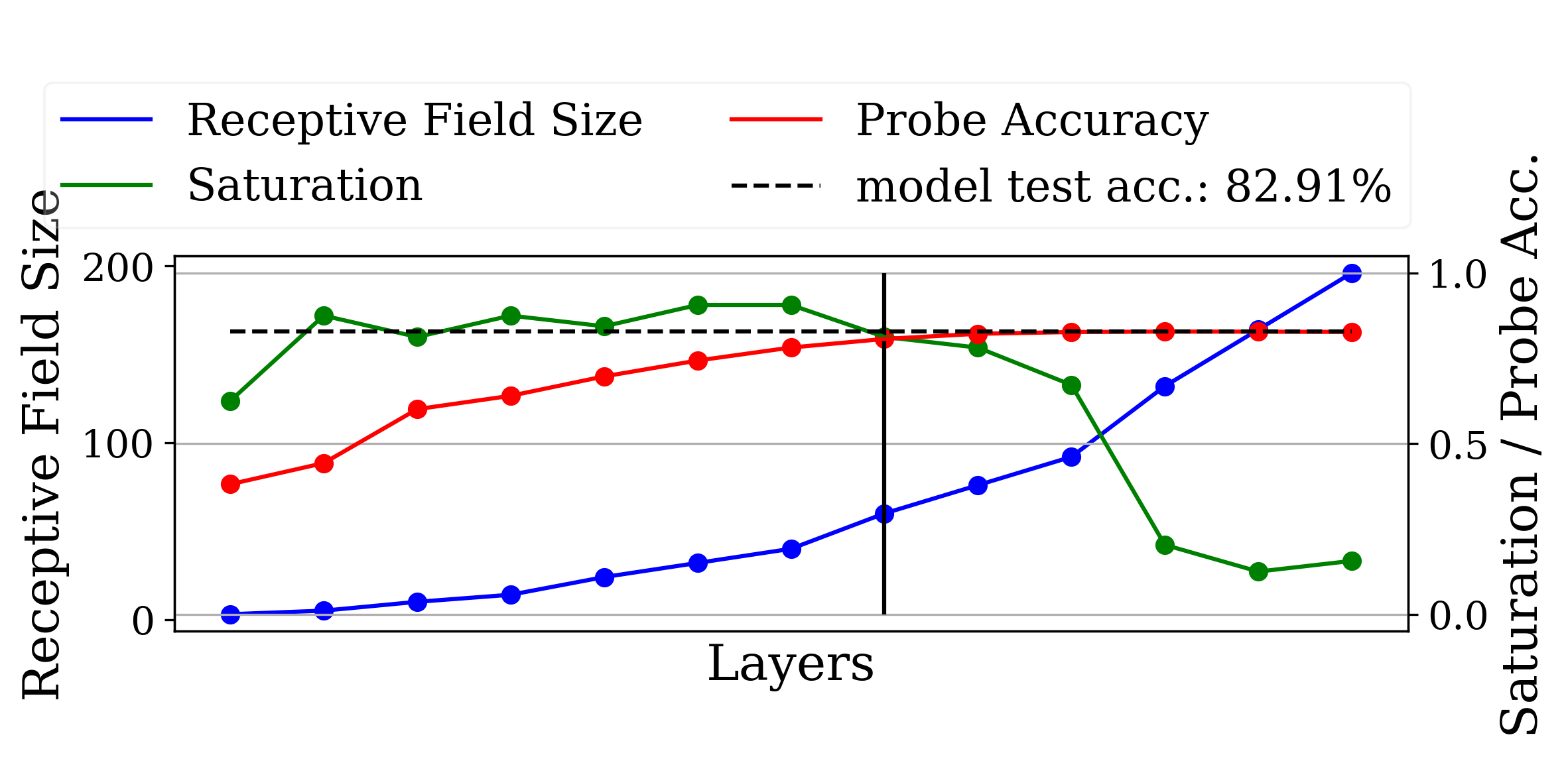}
	}
		\subfloat[VGG19]{
	    \includegraphics[width=0.4\columnwidth]{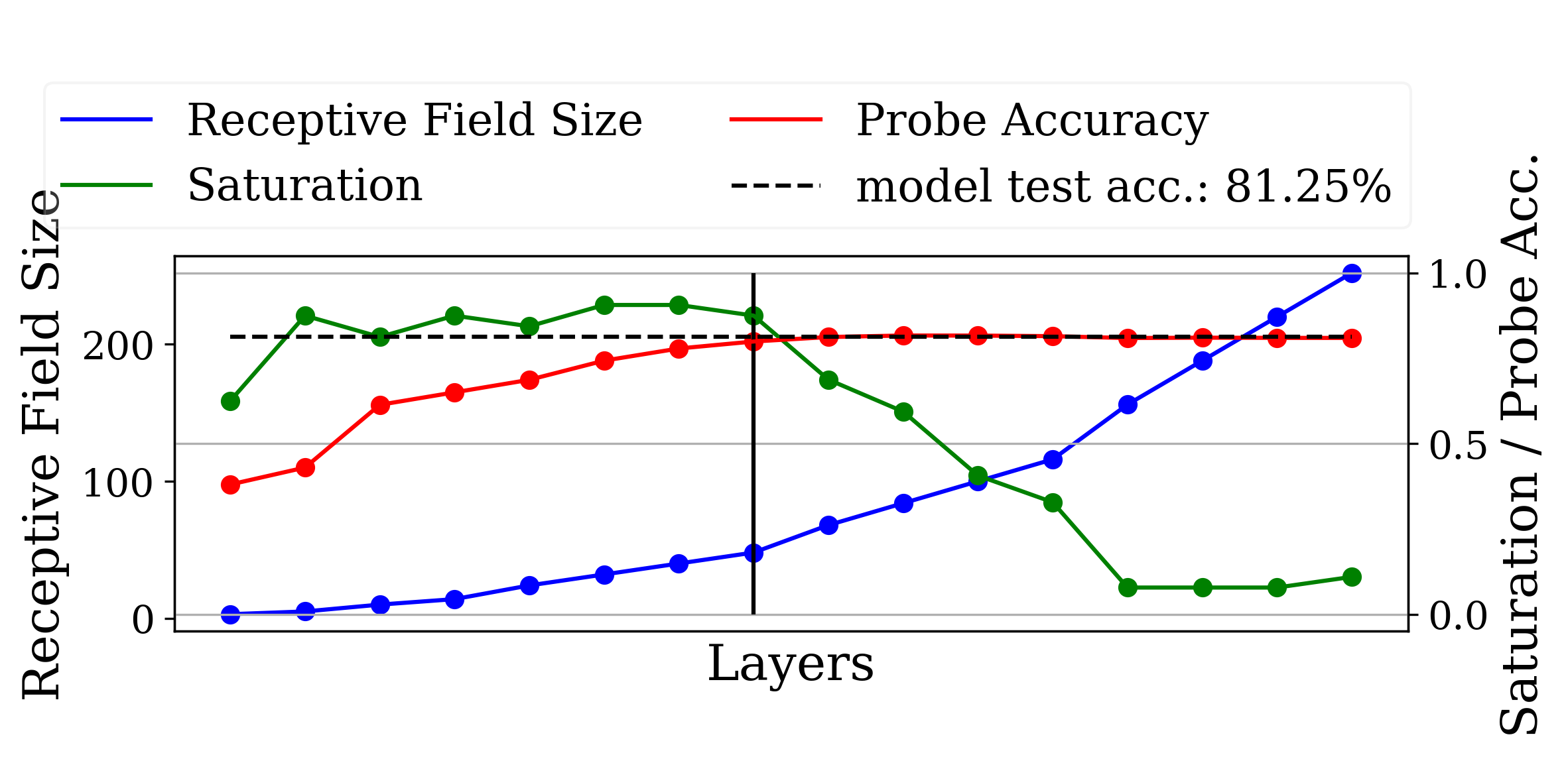}
	}
	\caption{Performance improvements of the logistic regression probes past the border layer are miniscule, even though the capacity of each layer is reduced to $\frac{1}{8}$ of the original capacity. This indicates that the networks are unable to shift processing to otherwise unused layers even if the capacity is limited. This is consistent with observation made by \ \cite{sizematters}.}
    \label{fig:border_layers}
\end{figure}

Even though the capacity of the networks has been significantly reduced in every layer, the networks do not spread the inference process among significantly more layers.

Based on these results we conclude that the inference dynamics of the tested networks did not change substantially by reducing their capacity.
This means that the capacity of layers primarily interacts with the difficulty of the problem, while the presence and absence of tail patterns interact with the receptive field as exemplified by \cite{sizematters}.

\section{Conclusion}
In this work, we explore the properties of saturation deeper and integrate this knowledge into insights by the authors of \cite{featurespace_saturation} and \cite{sizematters}.
We find that saturation is influenced in two ways.
First, while the saturation of layers relative to other layers in the network is indicative of mismatches between input resolution and architecture, the average saturation level provides insights into the interaction of the network's capacity and  the complexity of the problem. This observation opens another way to detect inefficiencies based on over- or under-saturation of the network.
Second, we find that these axes of analysis are independent, meaning that low capacity networks do not distribute their inference significantly different among layers than high capacity networks of the same architecture.
The observed saturation patterns converge similar to the loss of network and can thus be detected early during training allowing for quicker design iterations.
In summary, saturation can be useful as a tool for both, optimizing neural architectures for specific problems and for performing more principled investigation of information processing in deep neural networks.

\bibliographystyle{natbib}







\end{document}